\documentclass[10pt,twocolumn,letterpaper]{article}

\usepackage{ijcb}
\usepackage{times}
\usepackage{epsfig}
\usepackage{graphicx}
\usepackage{amsmath}
\usepackage{amssymb}
\usepackage[pagebackref]{hyperref}

\usepackage{tikz}
\usepackage{amsmath}
\usepackage{booktabs}

\ijcbfinalcopy 

\ifijcbfinal\fi
\begin{document}

\title{Towards Zero-Shot Interpretable Human Recognition: A 2D-3D Registration Framework}

\author{Henrique Jesus and Hugo Proença\\
IT - Instituto de Telecomunicações \\
University of Beira Interior, Portugal\\
{\tt\small henrique.jesus@ubi.pt, hugomcp@di.ubi.pt}}

\maketitle
\thispagestyle{empty}

\begin{abstract}
    Large vision models based in deep learning architectures have been consistently advancing the state-of-the-art in biometric recognition. However, three weaknesses are commonly reported for  such kind of approaches: 1) their extreme demands in terms of learning data; 2) the difficulties in generalising between different domains; and 3) the lack of interpretability/explainability, with biometrics being of particular interest, as it is important to provide evidence able to be used for forensics/legal purposes (e.g., in courts). 

    To the best of our knowledge, this paper describes the first recognition framework/strategy that  aims at addressing the three weaknesses simultaneously. At first, it relies exclusively in synthetic samples for learning purposes. Instead of requiring a large amount and variety of samples for each subject, the idea is to exclusively enroll a 3D point cloud per identity. Then, using generative strategies, we synthesize a very large (potentially infinite) number of samples, containing all the desired covariates (poses, clothing, distances, perspectives, lighting, occlusions,...). Upon the synthesizing method used, it is possible to adapt precisely to different kind of domains, which accounts for generalization purposes. Such data are then used to learn a model that performs local registration between image pairs, establishing positive correspondences between body parts that are the key, not only to recognition (according to cardinality and distribution), but also to provide an interpretable description of the response (e.g.: "both samples are from the same person, as they have similar facial shape, hair color and legs thickness").

\end{abstract}

\section{Introduction}

It is known that the remarkable ability of humans to recognize objects relies on prototypes somehow stored in our brain and matched to a compressed version of the input data, pre-processed by the visual system.  For example, when looking at a photograph of an individual 'A' whom we know, we can compare it with his 3D representation (prototype), justifying whether both correspond or not: \emph{"those are the arms and legs thickness and length of individual 'A', and the head shape is also the same"}. This process is not only the key for recognition, but also accounts for explainability by pointing out which parts of the image correspond to the prototype, as in a registration process.

\begin{figure}[t]
    \centering
    \includegraphics[width=\columnwidth]{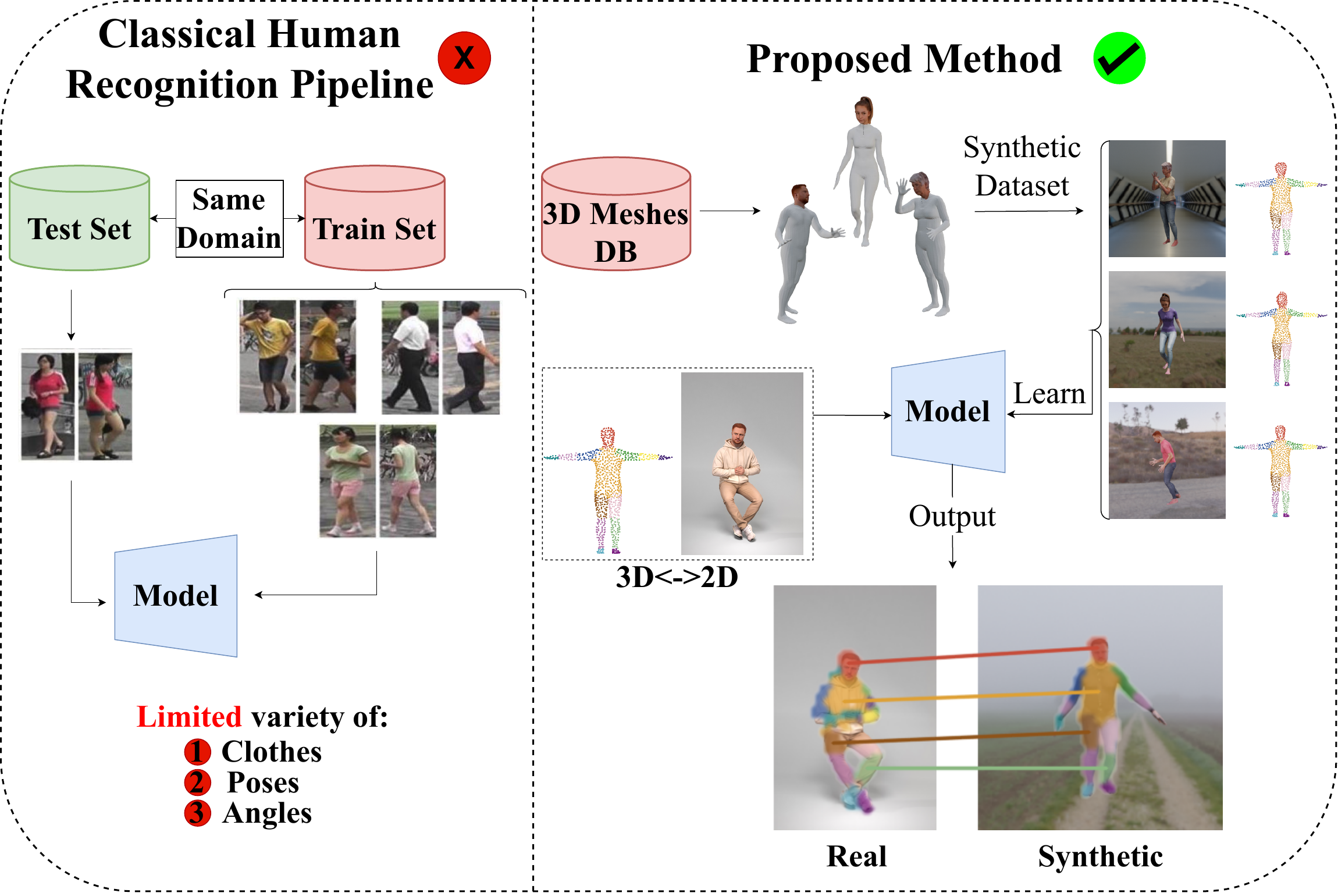}
    \caption{We propose an interpretable human recognition framework trained exclusively on synthetic data. In contrast to traditional methods that rely on datasets with limited variety of clothing, poses, and perspectives (e.g.,\cite{market}), our pipeline generated data with considerable variability. Our model learns by transferring knowledge from the subject in the image to a 3D representation of the same. In the end, it can perform recognition on real data, also providing human understandable explanations for the decisions taken, through registration.}
    \label{fig:esquema}
\end{figure}


While machine learning models excel at finding patterns and making predictions, their complex inner workings often remain a mystery. Legal requirements, such as the "right to explanation" in GDPR \cite{GDPR}, promote explainability in automated decision-making that significantly impacts individuals. This is where interpretability in machine learning becomes crucial, since it helps us comprehend how models make decisions. As a result, there's a growing emphasis on developing models that can justify the decisions made by neural networks. According to Zhang \etal \cite{interp_survey}, some models use techniques such as combining parts of the image with memorized prototypical features of an object, which is very similar to human mechanisms.

In this work, we present a framework able to provide interpretable cues about object recognition through  2D (image), 3D (prototype) registration. Fig. \ref{fig:esquema}, provides a cohesive perspective of the proposed framework. The model receives an image and a point cloud of an subject, aiming at learning semantic features from both modalities through knowledge transfer between them. If both representations regard the same person, it is expected that most correspondences align semantically, while the opposite should occur in case the subjects are different. However, in the latter case, if there are similar body parts between the image and the 3D object, the model should still associate them. Note that, under this paradigm, our focus is not exactly in creating a model with high  spatial registration performance, but rather infer the semantic interpretability through the identified correspondences between regions of the image and prototype data.

Biometric recognition in 2D images presents several challenges due to various well known factors, such as variations in clothing, poses, occlusions and ennvironmental features \cite{reident}. To overcome these challenges, the typical solution involves to collect extensive data for each individual, covering all the possible variation factors. However, this data collection and annotation process is highly time-consuming and labor-intensive, which substantially increases the costs associated to recognition. 

As a response to the above outlined difficulties, one of our constraints from the beginning was to rely exclusively in synthetic data for learning purposes. This allows to generate a practically unlimited number of images, capturing necessary variations without the logistical limitations associated to collecting extensive real data. Surely, even though we train the model using synthetic dataset, all the test experiments were carried in real data, highlighting its zero-shot learning capability \cite{zero_shot}. In our context, we use this term to express the model's ability to generalize to data modalities it hasn't seen during training. The rationale behind this lies in the capability to render high-quality data from just a 3D mesh of an individual. This enables the model to recognize individuals in entirely different conditions, such as real-world scenarios.\\

The remainder of this paper is divided into four sections: Section \ref{sec:related_work} presents the related work in the field of 2D-3D feature matching and interpretable object recognition. Section \ref{sec:proposed_method} describes in detail the proposed method. Section \ref{sec:experiments} addresses the experiments, obtained results, and their discussion. Finally, Section \ref{sec:conclusions} concludes the main key points of this work.

\section{Related Work}
\label{sec:related_work}

\subsection{2D-3D Feature Matching}

Before deep learning, handcrafted features dominated image registration tasks. These methods, relying on algorithms manually designed to detect image features, were crucial for aligning different views of the same scene. Essential characteristics of these features included robustness to changes in scale, rotation, and lighting. Among these, SIFT and ORB \cite{SIFT, ORB} stood out for their effectiveness and efficiency, setting a benchmark in the field of computer vision for image registration.

The emergence of Deep Learning and CNNs has allowed the transition from handcrafted to learned features in image feature extraction tasks. DeTone \etal \cite{Superpoint} presented a deep learning model that detects and describes keypoints in images. It is designed as a fully convolutional network, consisting of a shared encoder based on a VGG-style \cite{VGG} network, which reduces the dimensionality of the input image. Following this, the architecture splits into two separate decoder heads: one for detecting interest points and the other for describing these points.

For 2D-3D matching, Feng \etal \cite{matchnet} presents a triplet-like deep network architecture designed for matching keypoints between 2D images and 3D point clouds. The network has three branches: one for the 2D image keypoint and two for the 3D point cloud keypoint (negative and positive pair), with shared weights for the 3D branches. The image branch uses a VGG16 \cite{VGG} as feature extractor while the point cloud branches use PointNet \cite{pointnet}.
Most recent works, such as \cite{DeepI2p, CorrI2P} introduce different methods for aligning 2D images with 3D point clouds. In the first, the problem is treated like a classification task. In the latter, the method is responsible for converting the features of both image and point cloud into feature embeddings. Then, the overlapping areas between two modalities are identified using attention mechanisms.

\subsection{Semantic Correspondence}

Recent advancements in semantic correspondence have significantly improved the matching of similar objects and objects parts across images using deep learning techniques.

SFNet \cite{SFNet} uses binary foreground masks to guide a CNN in learning dense flow fields between image instances. The method emphasizes object-aware learning to reduce background clutter and employs a novel kernel soft argmax layer for accurate matching.

Deep ViT \cite{DeepViT} utilizes deep features from Vision Transformers (ViTs), specifically from a self-supervised DINO-ViT \cite{DINO-ViT} model. These features encode high-level semantic information and enable robust zero-shot applications in co-segmentation and semantic correspondences without extensive training, showing competitive performance against supervised methods.

Neural Best-Buddies \cite{BestBuddies} introduces a method for identifying sparse correspondences between images with semantically related parts. Utilizing a coarse-to-fine approach with deep features from a pre-trained CNN, this method finds mutual nearest neighbors at various abstraction levels and refines them to handle cross-domain variations.

\subsection{Interpretable Object Recognition}

According to Samek \etal \cite{interp_survey} we can say that in a general way interpretable methods are divided into two types: passive (\textit{post hoc}) and active. 

The \textit{post hoc} methods refer to the ability to interpret or explain the decisions made by a machine learning model after it has been trained \cite{interp_survey}. In this group, methods such as LIME and SHAP stand out \cite{LIME, SHAPY}. The former works by perturbing instances around a data point, observing original model predictions, and fitting a simpler model (e.g., linear regression) to explain the original model's behavior locally. The latter explains the model predictions by attributing the impact of each input feature, using principles from game theory. 

On the other hand, active models introduce pre-training modifications, such as additional network structures or adjustments to the training process, to enhance network interpretability \cite{interp_survey}. Inside this group, we can find the prototype-based methods. A prototype-based model is a method where instances are represented by a set of prototypes, typically examples from the dataset, which are used to interpret and classify new instances based on their similarity to these prototypes. Traditionally, methods such as those presented in Li \etal \cite{prototype_layer} introduced a prototype layer in a network to act as a prototype classifier. Predictions are determined by the proximity of inputs to the learned prototypes. Recent works, such as \cite{protopnet, int_obj_rec}, employ prototypes for interpretable object recognition, where these prototypes represent images of specific parts of an object. 

To the best of our knowledge, we are not aware of any prior work that utilizes parts of a 3D object as prototypes.

\section{Proposed Method}
\label{sec:proposed_method}
\begin{figure*}[ht] 
  \centering
  \includegraphics[width=\textwidth]{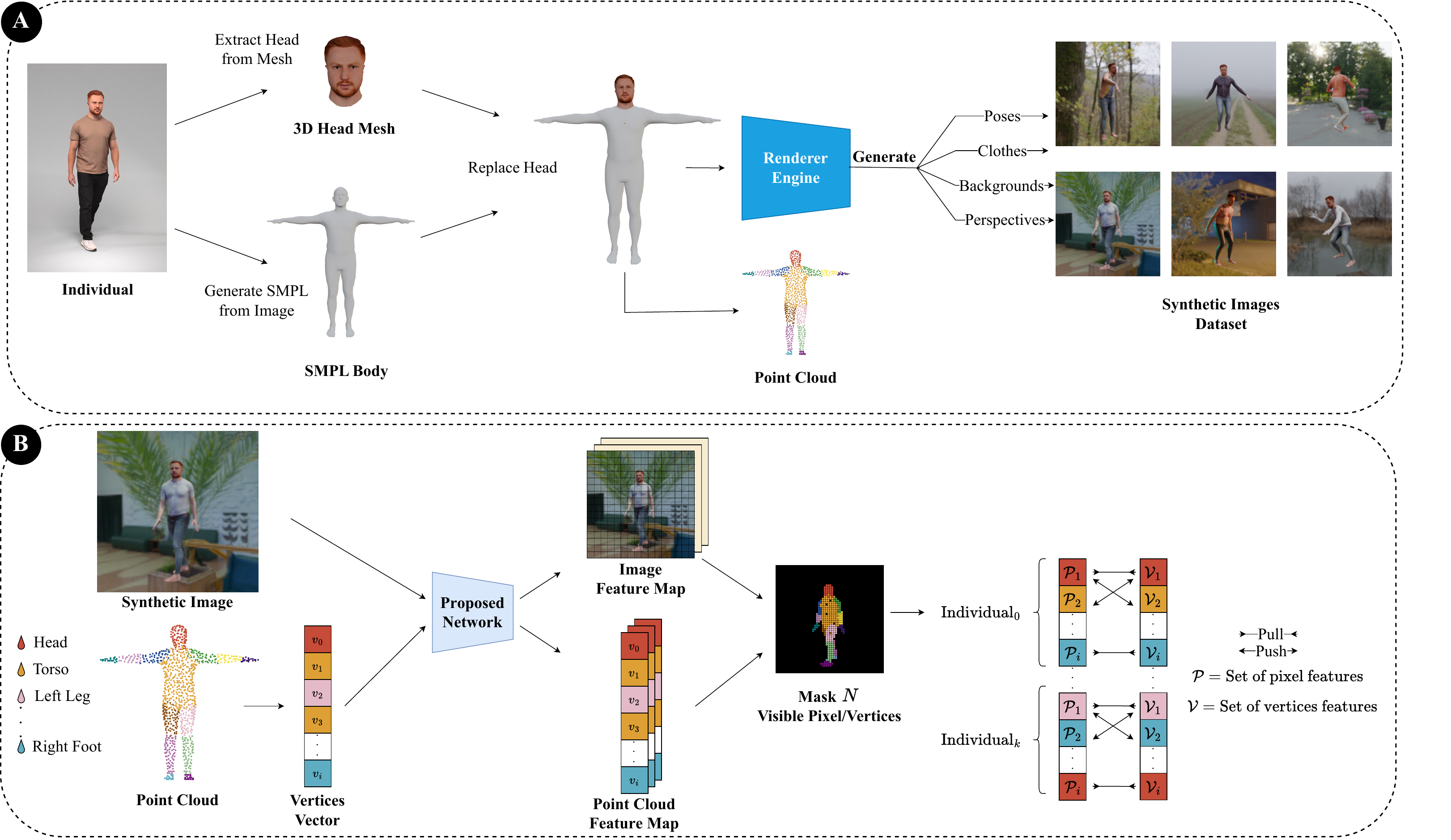}
  \caption{Cohesive perspective of the whole pipeline proposed in this paper: In the initial phase, we compile highly detailed 3D meshes of each individual and use them to generate images with a wide range of factors. In the subsequent phase, the model learns to semantically match the individual images with their 3D representation (point cloud) by pulling similar features together and pushing dissimilar features apart.}
  \label{fig:model_workflow}
\end{figure*}

Our proposed method is divided into three phases: generative phase, learning phase and inference phase. Fig. \ref{fig:model_workflow} illustrates the workflow of the generative and learning phases.

During the learning phase, the model aims to transfer the knowledge acquired from images of individuals to a 3D representation of the same, thus creating an interpretable prototype that can be used to semantically compare with other images. The generative phase is responsible for creating a large set of diverse images for a specific individual.

\subsection{Learning Phase}

Let $\mathbf{I}$ be an image with dimensions $3 \times H \times W$, and $\mathbf{P}$ a 3D object point cloud with dimensions $3 \times V$. Here, $H$ and $W$ represent the image's height and width, while $V$, set at 1024, represents the number of vertices. Thus, the proposed model takes a pair $\mathbf{I}$, $\mathbf{P}$ and returns: 1) a feature map of the image, $\mathbf{f_I}$, 2) a feature map of the point cloud, $\mathbf{f_P}$, and 3) a detector map, $\mathbf{d_I}$. $\mathbf{f_I}$ is a matrix with dimensions $H//4, W//4$, where each cell contains a feature vector characterizing semantically a region of pixels in $\mathbf{I}$. $\mathbf{d_I}$ is a binary mask that allows filtering the foreground from $\mathbf{f_I}$. $\mathbf{f_P}$ is a matrix with dimensions $V$, where each cell contains a feature vector characterizing semantically a vertex of $\mathbf{P}$.

The objective is to find the set of pixel features $\mathcal{P}$ in $\mathbf{f_I}$ and the set of vertex features $\mathcal{V}$ in $\mathbf{f_P}$ such that the distance between features of each element in $\mathcal{P}$ and $\mathcal{V}$ is minimal.

In order to semantically segment the prototype, we partition the point cloud of each individual into 14 parts, as illustrated in the diagram in Fig. \ref{fig:model_workflow} group "B". Since we use synthetic data, we can accurately obtain the projection matrix that maps the set of vertices of each part of the point cloud onto the image. This directly allows us to determine the segmentation of each region of the body in the image.

With this information, extracting $\mathcal{P}_i$ and $\mathcal{V}_i$ from $\mathbf{f_I}$ and $\mathbf{f_P}$, respectively, is trivial, where $i$ represents one of the 14 body parts. During training, the model is rewarded for pulling $\mathcal{P}_i$ close to the corresponding $\mathcal{V}_i$ and pushing the others away. For this purpose, we first concatenate these sets, such that $\mathcal{P} = \mathcal{P}_1 \oplus \mathcal{P}_2 \oplus \ldots \oplus \mathcal{P}_k$, i.e., $\mathcal{P} = \bigcup_{i=1}^{k} \mathcal{P}_i$, for the set of pixel features, and $\mathcal{V} = \bigcup_{i=1}^{k} \mathcal{V}_i$ for the set of vertex features. We do this for all individuals in the batch, concatenating all $\mathcal{P}$s into one, as well as the $\mathcal{V}$s, obtaining two large feature vectors, respectively $\mathcal{B}_P$ and $\mathcal{B}_V$. Then, we calculate the cosine similarity matrix $\mathbf{C}_\mathit{sim}$ between $\mathcal{B}_P$ and $\mathcal{B}_V$ as follows:

\begin{equation}
\label{eq:cosine-sim}
    \mathbf{C}_\mathit{sim} = \frac{ \mathcal{B}_P \cdot \mathcal{B}_V}{\|\mathcal{B}_P\| \|\mathcal{B}_V\|}.
\end{equation}

This similarity matrix is scaled by a learnable temperature parameter $\mathbf{t}$. Finally, we pass it through a sigmoid function and compute the binary cross entropy loss between $\mathbf{C}_\mathit{sim}$ and the ground truth correspondences.

\subsection{Inference Phase}

During inference, the process is somewhat similar to the learning phase. Here, we use $\mathbf{d}_\mathbf{I}$ to filter the foreground of $\mathbf{f}_\mathbf{I}$. The resulting features are compared with $\mathbf{f}_\mathbf{P}$ through cosine similarity and scaled by the parameter $\mathbf{t}$. The resulting matrix is passed through a sigmoid, and, finally, we filter the correspondences with confidence above a given threshold $\theta$. By organizing these correspondences with segmentation information, we obtain $\mathcal{P}$ and $\mathcal{V}$, indicating which pixels in the image $\mathbf{I}$ correspond semantically to the 3D representation $\mathbf{P}$.

\subsection{Generative Phase}

The pipeline used for our synthetic data generation method is illustrated in Fig. \ref{fig:model_workflow} in the "A" group. 

Firstly, we extract the SMPL \cite{SMPL} mesh of the individual using SMPLify \cite{SMPLify}. This type of mesh allows obtaining a 3D representation of the individual's body that is very close to their real physique. However, the SMPL \cite{SMPL} mesh does not preserve details of the individual's head, hair, and face. For this reason, we extract a detailed mesh of the head and replace the head of the SMPL \cite{SMPL} mesh with this one. In this way, we obtain a highly detailed and complete 3D representation of the individual. Further details on the extraction of the detailed mesh of head from each individual, and the process of replacing them on the 3D mesh generated through SMPL \cite{SMPL}, are described in the Subsection \ref{sub:implementation}.

To simulate the individual in different poses, we use the VPoser algorithm presented in \cite{VPoser}, which allows generating numerous coherent poses of an SMPL \cite{SMPL} mesh. Finally, we use the Blender rendering engine to generate various perspectives of the individual, alternating backgrounds, lighting, and clothing.

\subsection{Framework Architecture}

\begin{figure*}[ht]
    \centering
    \resizebox{0.6\linewidth}{!}{\input{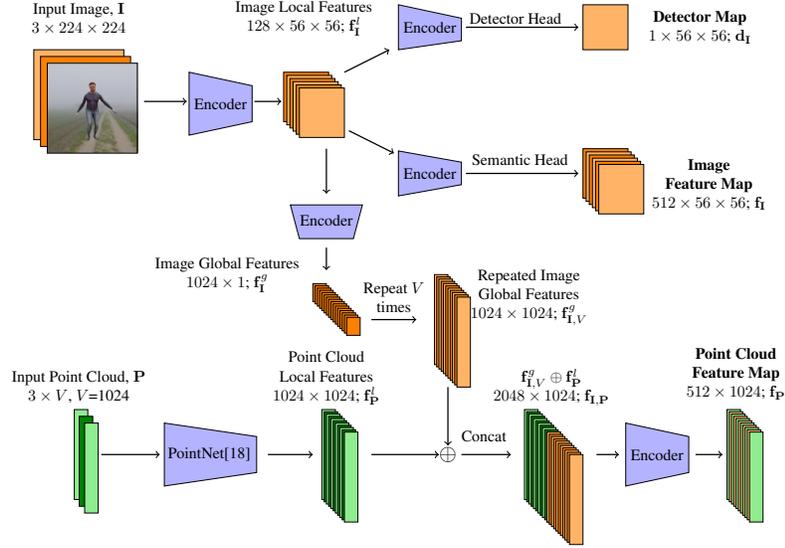}}
    \caption{Illustration of the proposed network architecture: the network divides into two branches, one responsible for the image and the other for the point cloud, with feature sharing between them. The image branch has two heads, one responsible for the detector map and the other for the feature map. In the point cloud branch, the features from the image are concatenated, and in the end, the point cloud feature map is returned.}
    \label{fig:model_arch}
\end{figure*}

The figure \ref{fig:model_arch} illustrates the architecture of our proposal model and it is divided into two main components: an image processing module and a 3D object processing module. The design facilitates the sharing of features between the two, allowing for a transfer of knowledge from the image to the 3D object.

The encoders in our model use a basic VGG \cite{VGG} style, with each block used in our network consisting primarily of a sequence of convolutional layers with a ($3\times1\times1$) configuration, followed by batch normalization and ReLU activation like in \cite{Superpoint}.

Initially, the input image $\mathbf{I}$ with dimensions $3\times(224\times224)$ is processed by an encoder aimed at extracting local image features $\mathbf{f}_{\mathbf{I}}^{l}$. This encoder utilizes sequences of 12 blocks to extract features from the image. It contains two Max Pooling layers to downsample the image by a factor of 2 each, resulting in local features with dimensions of $128\times(56\times56)$.

From this point, the network branches into two distinct heads: a detector head, and a semantic head, with a third encoder transforming the local features $\mathbf{f}_{\mathbf{I}}^{l}$ into global image features $\mathbf{f}_{\mathbf{I}}^{g}$. The detector head comprises just two blocks that convert $\mathbf{f}_{\mathbf{I}}^{l}$ into a binary detection map $\mathbf{d_I}$ with dimensions of $1\times(56\times56)$. The semantic head has 12 blocks and outputs the image feature map $\mathbf{f}_{\mathbf{I}}$ with dimensions of $256\times(56\times56)$.

For the 3D object processing, the point cloud of the object, with dimensions $3\times1024$, is passed through a PointNet \cite{pointnet} architecture, returning the local point cloud features $\mathbf{f}_{\mathbf{P}}^{l}$ with dimensions of $1024\times1024$.

To share features between the image and the 3D object, we first repeat $V$ times (1024) the global image features $\mathbf{f}_{\mathbf{I}}^{g}$, resulting in $\mathbf{f}_{\mathbf{I},V}^{g}$, and concatenate them with $\mathbf{f}_{\mathbf{P}}^{l}$ to obtain $\mathbf{f}_{\mathbf{I,P}}$ with dimensions of $2048\times1024$. This idea is similar to what Siyu Ren \etal did in their work \cite{CorrI2P}. This combined feature map then passes through a final encoder consisting of six blocks, producing the point cloud feature map $\mathbf{f}_{\mathbf{p}}$ with dimensions of $512\times1024$.

\section{Experiments and Discussion}
\label{sec:experiments}

In this section, we present the results obtained from our model after it was trained on the synthetic dataset and tested on the real images test set. We discuss the dataset rationale for 3D human generation and some implementation details. To perform a detailed analysis of the model's performance, we present and discuss both qualitative and quantitative results. Finally, we highlight some notable cases that are particularly worth mentioning.

\subsection{Datasets and Implementation Details}
\label{sub:implementation}

As mentioned earlier, a significant aspect of this work arises from the need for a large amount of data to train a model capable of human recognition. Many state-of-the-art datasets in this field, such as \cite{market}, provide ample data but with limited variability in factors. One of the main challenges is clothing, as the model should not rely on those to recognize individual characteristics. Other factors like poses and camera perspectives are also restricted in these datasets, justified by the practical difficulties in collecting diverse data that would satisfy such requirements. For this reason, we presented in \ref{sec:proposed_method} a method for generating synthetic data using 3D meshes of individuals. 

To the best of our knowledge, there is no specific dataset containing detailed 3D meshes of human physiognomies, free from clothing dependencies and with all the detail of the head region, a key factor humans use for recognition. RenderPeople \cite{Renderpeople} offers an extensive collection of high-quality 3D meshes of people, providing a dataset with 10 individuals (5 female and 5 male), each having 5 meshes with different poses and clothing. However, with only 5 poses and different outfits per individual, which is insufficient for our case, we cannot utilize the body mesh. Nevertheless, we can use the detailed head mesh, along with a coherent representation of each individual's physiognomy. To obtain 3D representations of each person in the dataset, we utilize SMPL \cite{SMPL} meshes, considered state-of-the-art for obtaining highly accurate 3D physiognomies. Using the algorithm proposed in \cite{SMPLify}, we regress a SMPL \cite{SMPL} mesh using real studio images provided by RenderPeople \cite{Renderpeople}. The resulting mesh is further enhanced with the anthropometric values of each person, also provided in the RenderPeople \cite{Renderpeople} dataset. The process of extracting and merging the head with the obtained meshes was manual and aided by Blender. For generating various poses, we use the VPoser \cite{VPoser} algorithm, compatible with SMPL \cite{SMPL} meshes, allowing for the creation of numerous coherent human poses. To simulate clothing, we used textures from \cite{texturas}, which offers a texture database compatible with SMPL \cite{SMPL}. To simulate environments with lighting, we use HDR backgrounds extracted from public repositories. In total, using Blender, we generate 25,000 images for each of the 10 individuals, with dimensions 224x224x3. For the test set, we use real studio images provided by RenderPeople \cite{Renderpeople}, totaling 50 images.

\subsection{Qualitative Results}

\begin{figure*}[t] 
  \centering
  \includegraphics[width=0.9\textwidth]{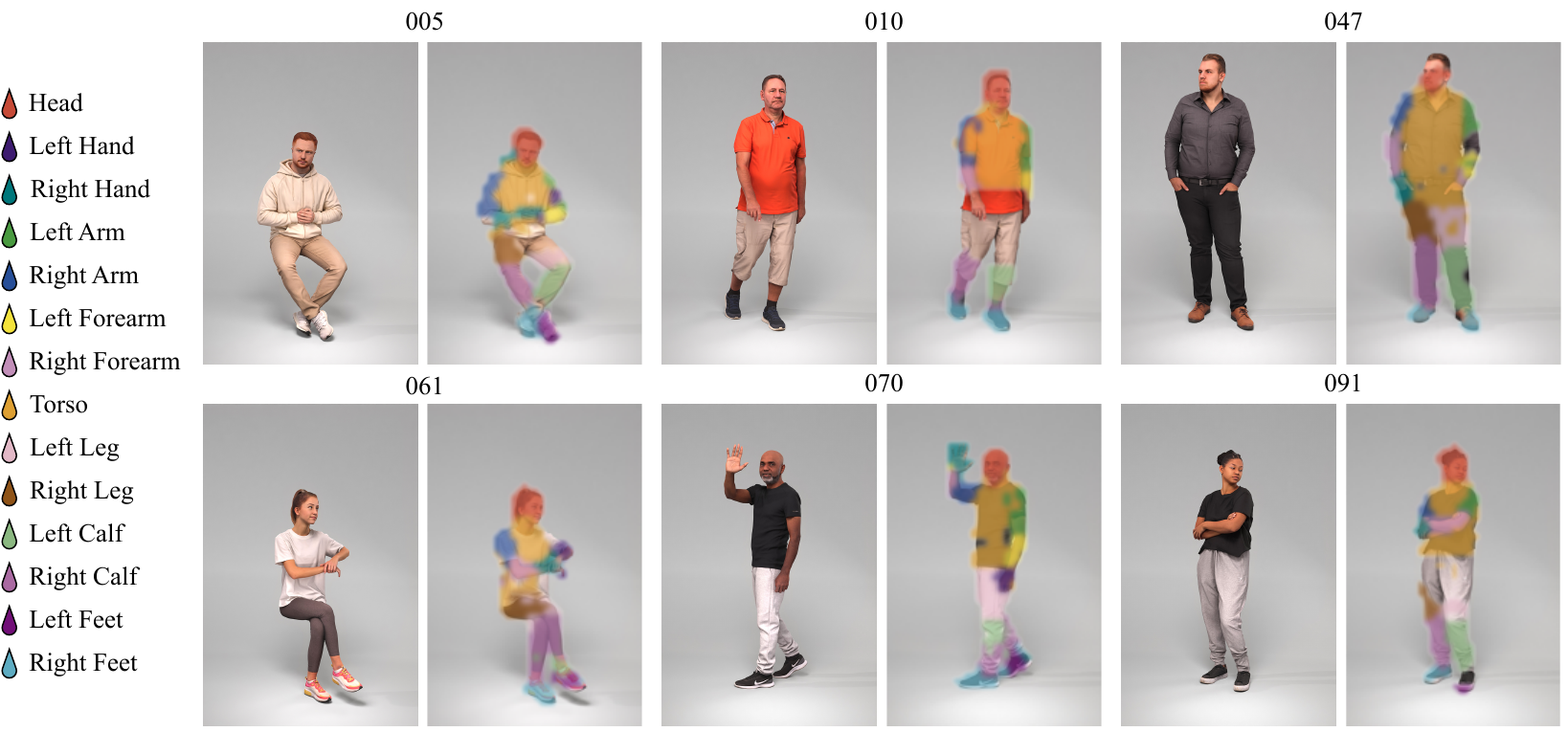}
  \caption{Examples of the results attained by our model. The colored regions represent the semantic matching between the individual in the image and the respective point cloud part.}
  \label{fig:qualitative_fig}
\end{figure*}

While the model was trained using synthetic data, it is more meaningful to evaluate its performance with real-world data. In this way, we tested the model with all the real images provided by RenderPeople \cite{Renderpeople}. For each of these images, the respective 3D representation was passed through. Figure \ref{fig:qualitative_fig} shows some examples of the obtained results, providing a general overview of the model's performance. Here, the colored regions symbolize the 2D-3D semantic registration, representing the region of image pixels that have semantic correspondences with the vertices of the point cloud.

Through qualitative analysis, we can conclude that the model demonstrates good performance in semantically associating individuals in the image with the 3D representation learned from synthetic data. This verifies the domain generalization capability of the model.

In the examples, we can observe that the regions of the head and torso were consistently associated correctly. Additionally, the upper and lower limbs were recognized correctly in most images. Occasionally, the model encounters some difficulty in detecting regions where the volume of clothing is greater, as seen in the images of individuals '010' and '091,' where a recognition gap around the waist is noticeable.

\subsection{Quantitative Results}

As previously mentioned, the model learns to semantically associate images with point clouds of the same individual. In other words, when this pair belongs to the same identity, is is expected numerous semantic correspondences, and the opposite when the pairs do not belong to the same identity. Consequently, we construct a confusion matrix based on the correspondence rate among all 10 identities. This approach allows us to assess the quantitative performance of the model in a different domain and analyze its generalization capabilities. However, given that the test images come from real images provided by RenderPeople \cite{Renderpeople}, it becomes impossible to directly compare the results obtained with other known methods. Since, the actual data from RenderPeople \cite{Renderpeople} lacks annotations for each body part segmentation, our objective is not to evaluate the performance of the correspondence distribution between each part. Instead, we aim to analyze the similarities distributions between each individual. To calculate the correspondence rate, $\rho$, we utilize the following formula:
\begin{equation}
    \rho = \sum_{i=1}^{14} \left( \frac{C_i}{C_{i_{\text{max}}}} \right),
\end{equation}
where $C_i$ refers the number of correspondences for each  part, and $C_{i_{max}}$ is the maximum number of possible correspondences for that part. 

\begin{figure*}[ht] 
  \centering
  \includegraphics[width=\textwidth]{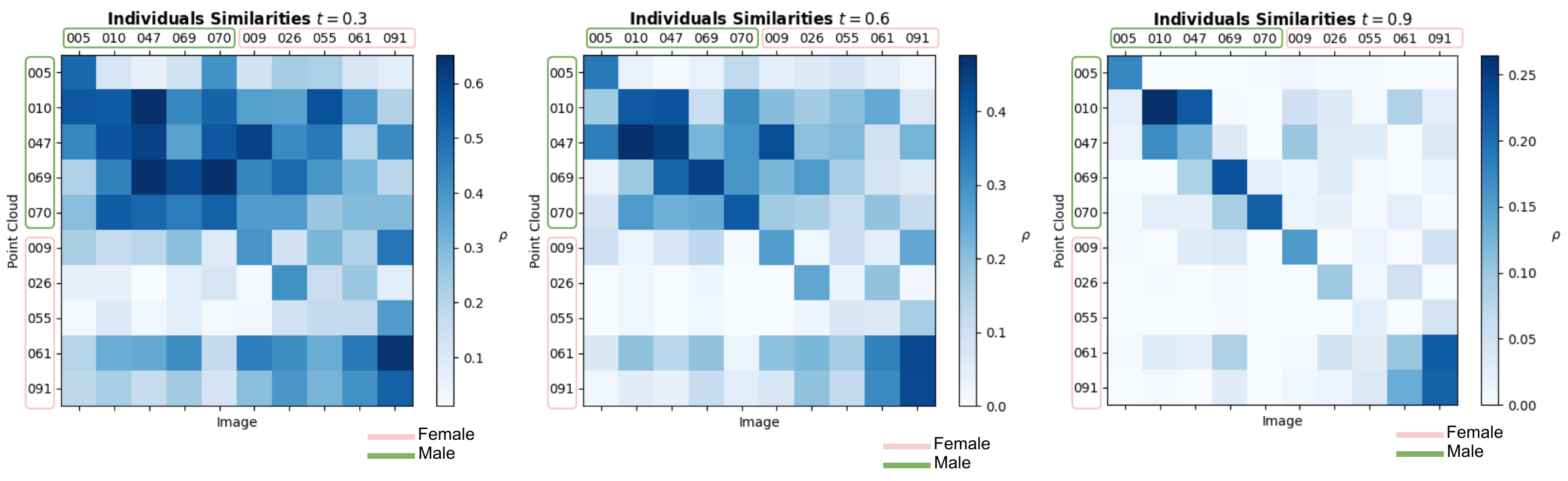}
  \caption{Matrix similarities between all individuals varying the threshold parameter $t$. The Y-axis represents the point clouds and X-axis the images. The scale is the correspondence rate $\rho$.}
  \label{fig:confusion_mat}
\end{figure*}

Fig. \ref{fig:confusion_mat} displays the confusion matrices of correspondence rates varying the threshold parameter, denoted as $t$. This parameter allows for adjusting the confidence of the obtained correspondences. We observe that a higher value of $t$ results in fewer false positives, emphasizing the diagonal elements of the matrices. Consequently, the maximum value of $\rho$ decreases as the quantity of correspondences also decreases. We can have a more objective idea by observing the F1-Scores of each class for $t=0.9$, by examining the Table \ref{tab:f1_scores}.

\begin{table}[!h]
\centering
\begin{tabular}{@{}lc@{}}
\toprule
\textbf{Class Name} & \textbf{F1-Score (0.9)} \\ \midrule
005 & 0.839 \\
010 & 0.434 \\
047 & 0.231 \\
069 & 0.502 \\
070 & 0.646 \\
009 & 0.478 \\
026 & 0.440 \\
055 & 0.220 \\
061 & 0.217 \\
091 & 0.406 \\ \bottomrule
\end{tabular}
\caption{F1-Scores per Class with $t=0.9$.}
\label{tab:f1_scores}
\end{table}

\begin{figure}[htbp]
  \centering
  \begin{minipage}{0.6\columnwidth}
    \centering
    \includegraphics[width=\linewidth]{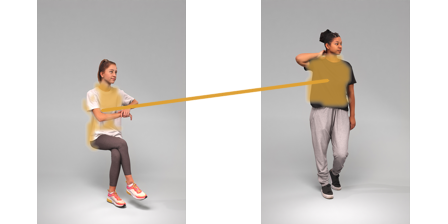}
  \end{minipage}\hfill
  \begin{minipage}{0.6\columnwidth}
    \centering
    \includegraphics[width=\linewidth]{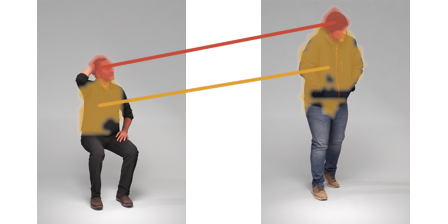}
  \end{minipage}
  \caption{Examples of the model mismatching regions in individuals. In the top image, a large shirt on the left confuses the model's association with the individual torso (yellow) on the right. In the bottom image, head (red) occlusion on the individual in the right adds to the model confusion since both have similar torso morphology.}
  \label{fig:mismatches}
\end{figure}

Therefore, the following observations will be made with an analysis of the confusion matrix at $t=0.9$. We observe that individuals '010' and '047' exhibit significant similarities to the extent that class '047' is entirely confused with '010'. Visually, these two individuals appear to have a very similar body morphology, both being endomorphic. Another factor contributing to this confusion by the model is that the test images of individual '047' exhibit occlusions in the head region. This makes it challenging for the model to semantically associate this region with the corresponding class. Also, the class '061' was confused with class '091'. In this case, both individuals have significantly different morphologies, with '061' having less body mass than '091'. However, in the test images of '061', the individual appears with large clothes, making it semantically inconsistent to associate it with a slimmer body. The factor that still allows it to have considerable similarities with its 3D representation is the head region. Fig. \ref{fig:mismatches} illustrates two examples in which the model semantically mismatches those individuals.

\begin{figure}[htbp]
  \centering
  \includegraphics[width=0.8\linewidth]{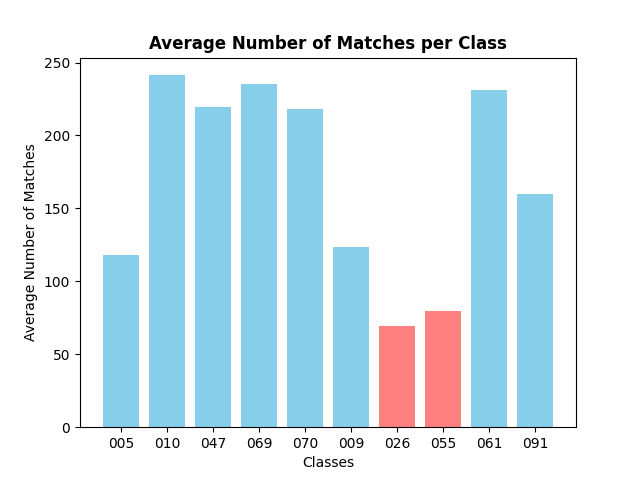}
  \caption{Average number of matches per class in the image feature map. The plot shows that class '026' and '055' were the ones with less number of correspondences on the test set.}
  \label{fig:number_of_matches}
\end{figure}

Regarding the class '055', this is the class where the model encountered the most difficulty in associating the corresponding point cloud. Fig. \ref{fig:number_of_matches} illustrates the average number of correspondences per class. Here, we can observe that indeed, there were fewer correspondences in the '026' and '055' classes. This can be attributed to the fact that the test images for these two classes feature clothing significantly different from the training set. Specifically, in some photos, individuals appear wearing long dresses, which confuses the model in semantically associating the body parts with the known 3D representation. 

As for the remaining classes, we can conclude that the model is able to recognize and semantically associate them with the correct 3D representation, demonstrating its domain generalization capability in these cases.

\subsection{Notable Cases}

As mentioned earlier, the model encounters some difficulties when the data varies significantly from the training set. The most notable cases where the model struggles are scenarios in which individuals' hair drastically alters the appearance of the face and when clothes seem to change the body's morphology.

\begin{figure}[htbp]
  \centering
  \includegraphics[width=0.7\linewidth]{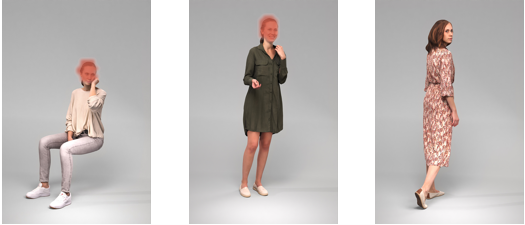}
  \caption{Notable examples where the model can recognize the individual's head (red) in the first two images but struggles in the third due to different hairstyles.}
  \label{fig:head_example}
\end{figure}

Fig. \ref{fig:head_example} illustrates a case where the model can semantically identify the head in the first two images, here the hair is tied. However, it fails in the last image because the loose hair omits some part of the head and neck.

\begin{figure}[htbp]
  \centering
  \includegraphics[width=0.5\linewidth]{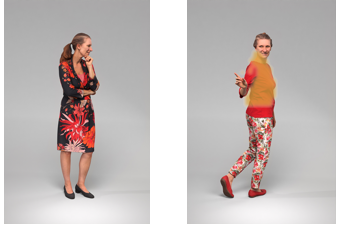}
  \caption{Notable example where the model can't recognize the individual's torso in the first image due to the long dress and the corresponding change in shape. However, the model performs perfectly in (yellow) the second image, since the subject is wearing a regular shirt.}
  \label{fig:no_corr}
\end{figure}

Another scenario is presented in Fig. \ref{fig:no_corr} here the model struggles to semantically associate the torso with the individual when the clothing significantly changes its morphology (left image). However, it successfully identifies the torso when the shirt has a more regular appearance (right image).

In further work, we could address the first case by adding more heads with different hairstyles to the mesh database, such as loose hair or tied hair. In the second case, instead of using 2D textures to simulate clothing, incorporating a database of clothing meshes could be considered. This way, the model could better generalize the body parts even when the volume of these parts differs from the original.

\section{Conclusions and Further Work}
\label{sec:conclusions}

In this paper, we presented a human recognition framework that learns exclusively from synthetic 3D data, but is able to work effectively in real-world and heterogenous domains scenarios, providing an interpretable description of its responses. We described a pipeline where we obtain a SMPL mesh \cite{SMPL} and modify it to accurately represent each subject body features. Upon this data, we can generate a potentially infinite learning set, with all the variability factors considered the most important (e.g., pose, clothing, distances, and lighting). The proposed model learns to transfer the semantic knowledge of each individual body parts in the images to 3D representations of the same individuals. 

Our experiments in real-world data revealed its significant domain generalization capability. Overall, the model can semantically associate individuals in the image with their 3D representation. We observed some challenges, primarily in images where individuals wear loose clothing, as it tends to confuse the model regarding body morphology. Another challenge observed was certain hairstyles that posed difficulties in matching the heads of some individuals.

Some ideas for future work involve using 3D meshes instead of textures to simulate clothing, providing the model with an even more accurate representation of real-world data. Another idea is to add more configurations of different hairstyles. Finally, implementing textual descriptions to justify semantic registration could be explored further.

\section*{Acknowledgements}

This work was funded by FCT/MEC through national funds and co-funded by FEDER - PT2020 partnership agreement under the projects UIDB/50008/2020, POCI-01-0247-FEDER- 033395.

{\small
\bibliographystyle{ieee}
\bibliography{Main}

\begin{thebibliography}{10}\itemsep=-1pt

\bibitem{BestBuddies}
K.~Aberman, J.~Liao, M.~Shi, D.~Lischinski, B.~Chen, and D.~Cohen-Or.
\newblock Neural best-buddies: sparse cross-domain correspondence.
\newblock {\em ACM Transactions on Graphics}, 37(4):1–14, July 2018.

\bibitem{DeepViT}
S.~Amir, Y.~Gandelsman, S.~Bagon, and T.~Dekel.
\newblock Deep vit features as dense visual descriptors, 2022.

\bibitem{SMPLify}
F.~Bogo, A.~Kanazawa, C.~Lassner, P.~Gehler, J.~Romero, and M.~J. Black.
\newblock Keep it smpl: Automatic estimation of 3d human pose and shape from a single image.
\newblock In {\em Computer Vision--ECCV 2016: 14th European Conference, Amsterdam, The Netherlands, October 11-14, 2016, Proceedings, Part V 14}, pages 561--578. Springer, 2016.

\bibitem{DINO-ViT}
M.~Caron, H.~Touvron, I.~Misra, H.~Jégou, J.~Mairal, P.~Bojanowski, and A.~Joulin.
\newblock Emerging properties in self-supervised vision transformers, 2021.

\bibitem{texturas}
D.~Casas and M.~Comino-Trinidad.
\newblock {SMPLitex: A Generative Model and Dataset for 3D Human Texture Estimation from Single Image}.
\newblock In {\em British Machine Vision Conference (BMVC)}, 2023.

\bibitem{protopnet}
C.~Chen, O.~Li, C.~Tao, A.~J. Barnett, J.~Su, and C.~Rudin.
\newblock This looks like that: Deep learning for interpretable image recognition, 2019.

\bibitem{Superpoint}
D.~DeTone, T.~Malisiewicz, and A.~Rabinovich.
\newblock Superpoint: Self-supervised interest point detection and description, 2018.

\bibitem{GDPR}
{European Parliament} and {Council of the European Union}.
\newblock Regulation ({EU}) 2016/679 of the {European} {Parliament} and of the {Council}.

\bibitem{matchnet}
M.~Feng, S.~Hu, M.~H. Ang, and G.~H. Lee.
\newblock 2d3d-matchnet: Learning to match keypoints across 2d image and 3d point cloud.
\newblock In {\em 2019 International Conference on Robotics and Automation (ICRA)}, pages 4790--4796. IEEE, 2019.

\bibitem{zero_shot}
C.~H. Lampert, H.~Nickisch, and S.~Harmeling.
\newblock Learning to detect unseen object classes by between-class attribute transfer.
\newblock In {\em 2009 IEEE conference on computer vision and pattern recognition}, pages 951--958. IEEE, 2009.

\bibitem{SFNet}
J.~Lee, D.~Kim, J.~Ponce, and B.~Ham.
\newblock Sfnet: Learning object-aware semantic correspondence, 2019.

\bibitem{DeepI2p}
J.~Li and G.~H. Lee.
\newblock Deepi2p: Image-to-point cloud registration via deep classification.
\newblock In {\em Proceedings of the IEEE/CVF Conference on Computer Vision and Pattern Recognition}, pages 15960--15969, 2021.

\bibitem{prototype_layer}
O.~Li, H.~Liu, C.~Chen, and C.~Rudin.
\newblock Deep learning for case-based reasoning through prototypes: A neural network that explains its predictions, 2017.

\bibitem{SMPL}
M.~Loper, N.~Mahmood, J.~Romero, G.~Pons-Moll, and M.~J. Black.
\newblock {SMPL}: A skinned multi-person linear model.
\newblock {\em ACM Trans. Graphics (Proc. SIGGRAPH Asia)}, 34(6):248:1--248:16, Oct. 2015.

\bibitem{SIFT}
D.~G. Lowe.
\newblock Object recognition from local scale-invariant features.
\newblock In {\em Proceedings of the seventh IEEE international conference on computer vision}, volume~2, pages 1150--1157. Ieee, 1999.

\bibitem{SHAPY}
S.~M. Lundberg and S.-I. Lee.
\newblock A unified approach to interpreting model predictions.
\newblock {\em Advances in neural information processing systems}, 30, 2017.

\bibitem{VPoser}
G.~Pavlakos, V.~Choutas, N.~Ghorbani, T.~Bolkart, A.~A.~A. Osman, D.~Tzionas, and M.~J. Black.
\newblock Expressive body capture: 3d hands, face, and body from a single image.
\newblock In {\em Proceedings IEEE Conf. on Computer Vision and Pattern Recognition (CVPR)}, 2019.

\bibitem{pointnet}
C.~R. Qi, H.~Su, K.~Mo, and L.~J. Guibas.
\newblock Pointnet: Deep learning on point sets for 3d classification and segmentation.
\newblock In {\em Proceedings of the IEEE conference on computer vision and pattern recognition}, pages 652--660, 2017.

\bibitem{CorrI2P}
S.~Ren, Y.~Zeng, J.~Hou, and X.~Chen.
\newblock Corri2p: Deep image-to-point cloud registration via dense correspondence.
\newblock {\em IEEE Transactions on Circuits and Systems for Video Technology}, 33(3):1198--1208, 2022.

\bibitem{Renderpeople}
{RenderPeople}.
\newblock Renderpeople website.
\newblock \url{https://renderpeople.com/}, Nov 2023.

\bibitem{LIME}
M.~T. Ribeiro, S.~Singh, and C.~Guestrin.
\newblock " why should i trust you?" explaining the predictions of any classifier.
\newblock In {\em Proceedings of the 22nd ACM SIGKDD international conference on knowledge discovery and data mining}, pages 1135--1144, 2016.

\bibitem{ORB}
E.~Rublee, V.~Rabaud, K.~Konolige, and G.~Bradski.
\newblock Orb: An efficient alternative to sift or surf.
\newblock In {\em 2011 International Conference on Computer Vision}, pages 2564--2571, 2011.

\bibitem{VGG}
K.~Simonyan and A.~Zisserman.
\newblock Very deep convolutional networks for large-scale image recognition.
\newblock {\em arXiv preprint arXiv:1409.1556}, 2014.

\bibitem{int_obj_rec}
Q.~Wan, R.~Wang, and X.~Chen.
\newblock Interpretable object recognition by semantic prototype analysis.
\newblock In {\em Proceedings of the IEEE/CVF Winter Conference on Applications of Computer Vision}, pages 800--809, 2024.

\bibitem{reident}
M.~Ye, J.~Shen, G.~Lin, T.~Xiang, L.~Shao, and S.~C. Hoi.
\newblock Deep learning for person re-identification: A survey and outlook.
\newblock {\em IEEE transactions on pattern analysis and machine intelligence}, 44(6):2872--2893, 2021.

\bibitem{interp_survey}
Y.~Zhang, P.~Ti{\v{n}}o, A.~Leonardis, and K.~Tang.
\newblock A survey on neural network interpretability.
\newblock {\em IEEE Transactions on Emerging Topics in Computational Intelligence}, 5(5):726--742, 2021.

\bibitem{market}
L.~Zheng, L.~Shen, L.~Tian, S.~Wang, J.~Wang, and Q.~Tian.
\newblock Scalable person re-identification: A benchmark.
\newblock In {\em Computer Vision, IEEE International Conference on}, 2015.

\end{thebibliography}
}

\end{document}